\documentclass[10pt,journal,compsoc]{IEEEtran}

\ifCLASSOPTIONcompsoc
  \usepackage[compress]{cite}
\else
  \usepackage{cite}
\fi
\ifCLASSINFOpdf
\else
\fi
\usepackage{times}
\usepackage{soul}
\usepackage{url}
\usepackage[hidelinks]{hyperref}
\usepackage[utf8]{inputenc}
\usepackage[small]{caption}
\usepackage{graphicx}
\usepackage{amsmath}
\usepackage{amsthm}
\usepackage{booktabs}
\usepackage{algorithm}
\usepackage{algorithmic}
\usepackage[switch]{lineno}
\usepackage{amsmath,bm}
\usepackage{amssymb}
\usepackage{enumitem}

\begin{document}
%
\title{A Survey on Incomplete Multi-label Learning: \\ Recent Advances and Future Trends}
%
%
%
%
\author{Xiang Li, 
        Jiexi Liu,
        Xinrui Wang,
        and Songcan Chen$^{\dag}$,~\IEEEmembership{Senior Member,~IEEE}
        
\IEEEcompsocitemizethanks{\IEEEcompsocthanksitem Songcan Chen is with the College of Computer Science and Technology, Nanjing University of Aeronautics and Astronautics, and also with MIIT Key Laboratory of Pattern Analysis and Machine Intelligence.\protect\\
The corresponding author is Songcan Chen.

E-mail: s.chen@nuaa.edu.cn
\IEEEcompsocthanksitem Xiang Li, Jiexi Liu and Xinrui Wang are with the College of Computer Science and Technology, Nanjing University of Aeronautics and Astronautics, and also with MIIT Key Laboratory of Pattern Analysis and Machine Intelligence.}
\thanks{Manuscript received April 19, 2005; revised August 26, 2015.}}

%
%

\markboth{Journal of \LaTeX\ Class Files,~Vol.~14, No.~8, August~2015}%
{Shell \MakeLowercase{\textit{et al.}}: Bare Advanced Demo of IEEEtran.cls for IEEE Computer Society Journals}
%



\IEEEtitleabstractindextext{%
\begin{abstract}
In reality, data often exhibit associations with multiple labels, making multi-label learning (MLL) become a prominent research topic. The last two decades have witnessed the success of MLL, which is indispensable from complete and accurate supervised information. However, obtaining such information in practice is always laborious and sometimes even impossible. To circumvent this dilemma, incomplete multi-label learning (InMLL) has emerged, aiming to learn from incomplete labeled data. To date, enormous InMLL works have been proposed to narrow the performance gap with complete MLL, whereas a systematic review for InMLL is still absent. In this paper, we not only attempt to fill the lacuna but also strive to pave the way for innovative research. Specifically, we retrospect the origin of InMLL, analyze the challenges of InMLL, and make a taxonomy of InMLL from the data-oriented and algorithm-oriented perspectives, respectively. Besides, we also present real applications of InMLL in various domains. More importantly, we highlight several potential future trends, including four open problems that are more in line with practice and three under-explored/unexplored techniques in addressing the challenges of InMLL, which may shed new light on developing novel research directions in the field of InMLL.
\end{abstract}

}

\maketitle

\IEEEdisplaynontitleabstractindextext

%
\IEEEpeerreviewmaketitle

\ifCLASSOPTIONcompsoc
\IEEEraisesectionheading{\section{Introduction}\label{sec:introduction}}
\else
\section{Introduction}
\label{sec:introduction}
\fi

%
%
%
%
\IEEEPARstart{T}{he} co-occurrence nature of multi-objects or the hierarchical nature of a single object makes the real-world manifest in a multi-label form. For instance, for the former, an image may simultaneously contain the objects of mountain, tree, sun, and cloud; a document may concurrently cover the topics of artificial intelligence, machine learning, and deep learning. For the latter, a piece of clothing may be labeled as men, tops, and T-shirts; an apple tree can belong to the classes of Plant, Angiosperms, and Malus domestica at the same time.
Assigning the multiple associated labels to the corresponding sample has arisen the emergence of multi-label learning (MLL) \cite{sorower2010literature,zhang2013review,liu2021emerging}. The preceding two decades have witnessed its success in various applications, including image classification \cite{wei2015hcp,wang2016cnn,zhu2017learning,lanchantin2021general,zhang2023spatial,guo2023texts,zhu2023scene,ma2024target}, text categorization \cite{nam2014large,liu2017deep,xiao2019label,huang2019hierarchical,chang2020taming,zhang2021correlation,wiegmann2023trigger,chai2024compositional}, video analysis \cite{xu2011ensemble,xu2012semi,markatopoulou2018implicit,ray2018scenes,lu2021learning,tirupattur2021modeling,gupta2023class,guo2024satellite}, and medical science \cite{qiao2020mhm,liu2021effective,yogarajan2021transformers,reiss2021every,zhang2023triplet,jeong2023optimized,lai2023single,mahapatra2024gandalf}, just to name a few.

\begin{figure}[t]
	\centering
	\includegraphics[width=0.97\linewidth]{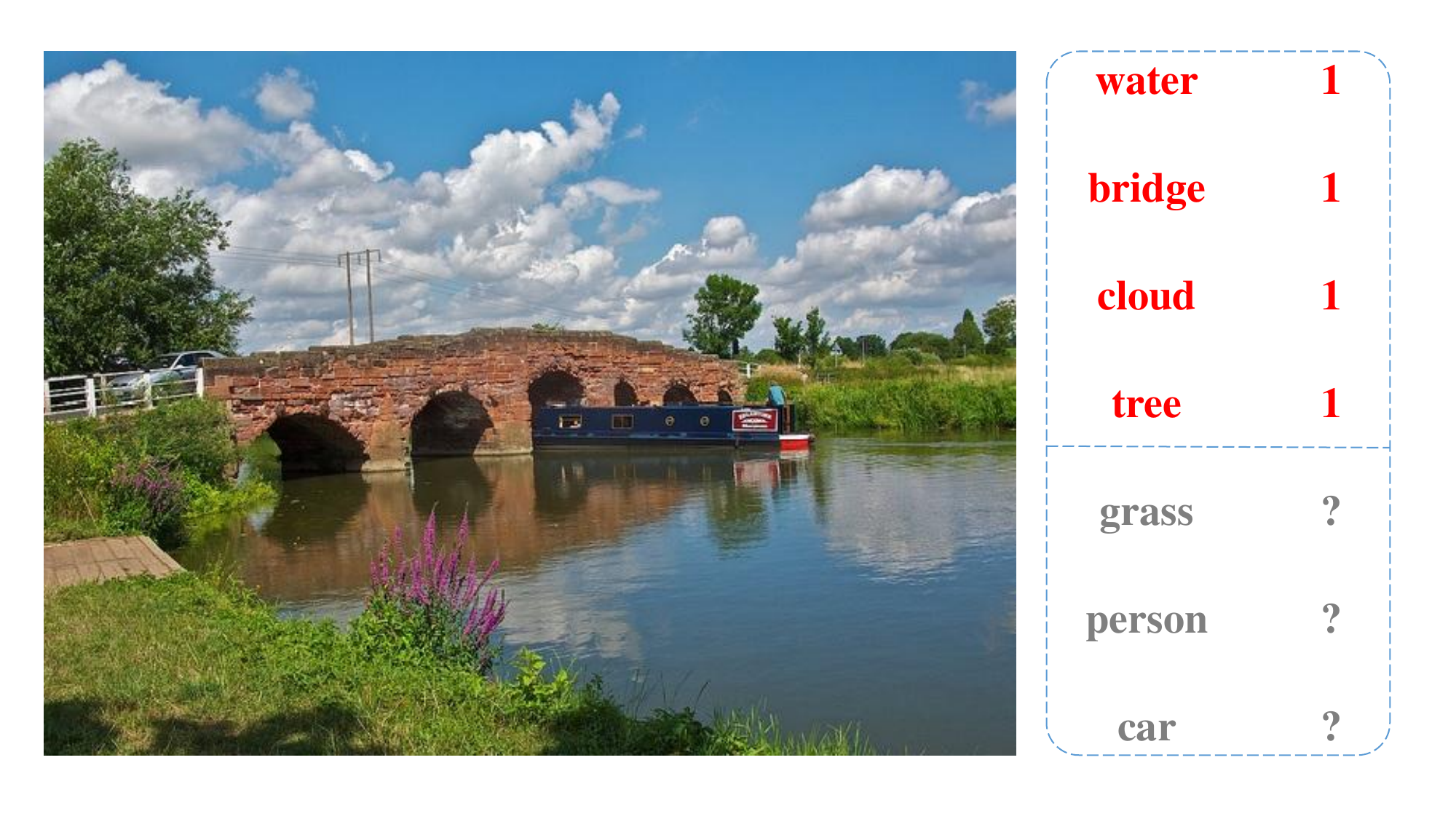} 
	\caption{An illustration of incomplete multi-label data. The left part is an image from real-world and labels in the upper right with red color and ``1" are the given labels, and labels in the lower right with gray color and ``?" are the missing labels. Note that all the seven labels are in the ground truth.}
	\label{fig1}
\end{figure}

The success of MLL is indispensable from the availability of complete and accurate supervision information\cite{lin2022rethinking,zhao2015semi}. However, acquiring such information is not only time-consuming and labor-intensive but sometimes also impossible in practice due to the inherent complexity and limited knowledge. To alleviate this dilemma, the paradigm of incomplete multi-label learning (InMLL) has emerged\cite{liu2006semi,sun2010multi}, where only a subset of labels needs to be provided for the dataset, and the goal is to learn from the incomplete or missing labeled data. Figure \ref{fig1} depicts an illustration of incomplete multi-label data.

The main reasons for the emergence of incomplete multi-label learning lie in two aspects. First, the low quality of dataset itself may cause the incompleteness. For instance, in an image, insufficient lighting or low resolution may render some blurry or tiny targets indistinguishable; in an audio, the intertwining of noisy background sounds with multiple target sounds may interfere with and influence each other, making them difficult to differentiate. Second, limited knowledge of the annotators will also produce incomplete labeled data. For example, annotators lacking specialized medical knowledge are prone to issues such as omission or mislabeling in medical image annotation, since their knowledge mismatches with the dataset being annotated. 

Note that obtaining data with incomplete labels is relatively simple and feasible, meeting the demand for environmental friendly and resource-conserving practices. Consequently, numerous researchers are dedicated into InMLL, aiming to achieve satisfactory performance while reducing labeling costs.

The inception of InMLL can be traced back to 2006, as evidenced by the pioneering work \cite{liu2006semi} on semi-supervised MLL. Subsequently, in 2010, authors of literature \cite{sun2010multi} proposed the first weakly supervised MLL work. The allure of this learning paradigm lies its the ability to significantly reduce labeling costs, enhancing its applicability and thereby attracting considerable attention and extensive research over the past years. Recognizing that the performance of InMLL is typically inferior to that of complete MLL due to the insufficiency of supervised information, enormous works in InMLL \cite{bucak2011multi,yu2014large,yang2016improving,jain2017scalable,durand2019learning,huynh2020interactive,cole2021multi,li2021concise,dong2022revisiting,chen2023semantic,wang2024missing,liu2024masked} have been proposed to narrow the performance gap between these two paradigms.

Despite the progress achieved in these works, there still lacks a comprehensive review that systematically summarizes the current progress and identifies unresolved challenges. Note that, even the latest MLL survey \cite{tarekegn2024deep} only regards InMLL as an open challenge and overlooks the progress of this field over the past decade. Thus, there is a timely need for a comprehensive survey in the InMLL field. In this paper, by thoroughly reviewing existing methods coupled with an analysis of their strengths and limitations, we not only attempt to fill the lacuna but also strive to pave the way for innovative research. Specifically, we present the formal definition of InMLL in Section \ref{sec2}. Moreover, we list six challenges of InMLL in Section \ref{sec3n}. Afterwards, we make a taxonomy of InMLL from the data-oriented and algorithm-oriented perspectives in Section \ref{sec3}. Besides, we also present several real applications for InMLL in Section \ref{sec4}. Finally, in Section \ref{sec6}, we point out four and three potential future directions from the perspectives of problem setting and technical route, respectively.


 
\section{The Definition of InMLL} \label{sec2}
In this section, we present the formal definition of InMLL. InMLL is a variant of MLL where the training data contains samples with only a subset of the possible labels available, rather than a complete label set. Formally, it can be defined as follows:

Let $\mathcal{X} \in \mathbb{R}^d$ be the $d$-dimensional feature space and $\mathcal{Y}=\{y_1,y_2,\cdots, y_C \}$ be the label space with $C$ possible class labels. Given a multi-label training set $\mathcal{D} = \{\mathbf{X}_i, \mathbf{Y}_i\}_{i=1}^N$, where $N$ is the number of training samples, $\mathbf{X}_i \in \mathcal{X}$ and $\mathbf{Y}_i \in \{1, -1, ?\}^C$ are the feature vector and the incomplete label vector corresponding to the $i$-th sample. Here, $\mathbf{Y}_{ij}=1$ ($j \in \{1,2,\cdots,C\}$) indicates that the $j$-th label is relevant to the $i$-th sample, $\mathbf{Y}_{ij}=-1$ indicates it is irrelevant, and $\mathbf{Y}_{ij}=?$ indicates that the relevance of the $j$-th label to the $i$-th sample is unknown or missing. The objective of InMLL is to learn a function $f:\mathcal{X} \longmapsto 2^\mathcal{Y}$ from the incompletely labeled training set $\mathcal{D}$ for predicting the labels of test samples. 

To facilitate understanding of this definition, we provide an illustrative example below. Without loss of generality, suppose $C=5$, and the label vector of the $i$-th sample $\mathbf{Y}_i$ is $[1,-1,1,?,?]$, where ``$1$" denotes that the sample contains the first and third labels, also known as positive labels, ``$-1$" denotes that the sample does not contain the second label, referred to as negative label, while ``$?$" denotes the values of fourth and fifth labels are missing, implying uncertainty regarding whether these labels are present in the sample. Given the label vector $\mathbf{Y}_i$, its corresponding indicator vector $\mathbf{\Omega}_i$ is $[1,1,1,0,0]$, where ``$1$" denotes the labels are provided, whereas ``$0$" denotes that the labels are missing. 

\section{Challenges of InMLL} \label{sec3n}
The challenges of InMLL primarily lie in the following aspects: the insufficiency of the supervised information, the inconsistency of label correlations, the serious imbalance of labels, and the presence of noisy labels. Besides, when designing algorithms, two common challenges are frequently encountered.

\subsection{The Primary Challenge}

\noindent{\textbf{The insufficiency of supervised information.} In the context of InMLL, labels are incomplete or missing, resulting in a quite limited supervised information. Consequently, the discriminative information guiding for training is insufficient, leading to the performance of InMLL degenerates significantly compared with that of complete MLL. This constitutes the primary challenge of InMLL and also serves as the main reason for its performance degeneration. To tackle this challenge, numerous InMLL works \cite{bi2014multilabel,xu2014learning,kong2014large,jain2016extreme,zhu2017multi,akbarnejad2018efficient,braytee2019correlated,ma2019label,ibrahim2020confidence,ma2021expand,schultheis2022missing,zhang2023learning,liu2024attention,tan2024two,yin2024feature} have made endeavor, which further brings the following challenges.

\subsection{Three Specific Challenges}
	
\noindent{\textbf{The inconsistency of label correlations.} Label correlations remain at the core of multi-label learning, whether within the context of complete or incomplete settings. Many InMLL works \cite{xu2014learning,yang2016improving,zhu2017multi,nguyen2019multi,feng2020regularized,gerych2022robust,wang2024missing,jia2024discriminative} have delved into exploring label correlations to ease the insufficiency of labels. However, the missing labels may cause the inconsistency of label correlations. For instance, the labels of ``cloud" and ``sky" often co-occur, resulting in a generally high correlation between them, while in the InMLL scenario, one of the ``cloud" and ``sky" labels may be missing, in this case, the correlation between them becomes low, which is inconsistent with the label correlation in the complete MLL scenario. 

\noindent{\textbf{The serious imbalance of labels.} Label imbalance is inherent in MLL \cite{wu2016constrained,braytee2019correlated,zhang2020towards,duarte2021plm,tarekegn2021review,ben2022multi,zhang2023learning,song2024toward}, characterized by two distinct types. The first is the imbalance between the positive and negative labels. Generally, labels in MLL are often sparse, making the number of negative labels far more than that of positive labels. To ease the insufficiency of labels, the missing labels are often assumed to be negative labels in the context of InMLL, which undoubtedly exacerbates the imbalance between the positive and negative labels. The second is the imbalance among the positive labels. Intuitively, different objects have different chances to appearing. Common objects always have a greater chance than the rare objects to appearing, which definitely causes the imbalance among them. In the context of InMLL, the insufficiency of the labels of rare objects will make such imbalance more serious.
	
\noindent{\textbf{The presence of noisy labels.} In MLL field, due to that the number of negative labels is far more than that of positive labels, the missing labels are often imputed as negative labels, which inevitably producing false negatives as noisy labels. However, existing works often neglect this issue and utilize the commonly used loss function, such as binary cross entropy (BCE) \cite{cole2021multi}, focal loss \cite{lin2017focal}, and asymmetric loss \cite{ridnik2021asymmetric} for training. Unfortunately, none of these losses are capable of dealing with the challenge of noisy labels, which also hinders the performance of InMLL.

\subsection{Two Common Challenges}
	
In addition to the above four challenges, when designing specific InMLL algorithms, the following two common challenges will also be encountered. 

\noindent{\textbf{What assumptions should be made?} To alleviate the issue of insufficient supervised information stemming from missing labels, InMLL algorithms need to introduce inductive biases or make various assumptions. However, in real-world, the true distribution of data is generally unknown. Consequently, the challenge arises in determining what assumption should be made to align with the latent data distribution.
	
\noindent{\textbf{How to reduce the costs of model selection?} Given the dramatic performance degradation under the InMLL scenario, InMLL algorithms often resort to adopting multiple assumptions to combat such a decline. However, multiple assumptions typically accompanied by multiple regularization terms and their corresponding hyper-parameters. When selecting the optimal model, multiple parameters will affect each other, and with each additional parameter, the difficulty and computational complexity of choosing the optimal parameters will substantially increase. Therefore, the challenge lies in determining how to make as few yet consistent assumptions as possible, tailored for the problems at hand, to reduce the costs associated with model selection substantially.

In summary, improving the performance of InMLL algorithms hinges predominantly on addressing challenge of insufficient supervised information. Researchers should be careful when utilizing the label correlations, and the challenges of imbalanced and noisy labels should be emphasized. Besides, what assumptions to adopt and how to reduce the cost of model selection should also be considered for designing an effective and efficient algorithm.

\section{A Taxonomy of InMLL} \label{sec3}
In this section, we make a taxonomy of InMLL from the data-oriented and algorithm-oriented perspectives, respectively.

\subsection{Data-Oriented}

From the data-oriented perspective, we classify InMLL works into two categories based on the label type and label missing type. More precisely, according to the label type, InMLL works can be divided into two subcategories: incomplete hard multi-label learning and incomplete soft multi-label learning.

\noindent{\textbf{Incomplete hard multi-label learning}} When the label vector is defined as a $C$-dimensional integer vector with missing values, then learning from such incomplete data is called incomplete hard multi-label learning. Formally, given a multi-label training set $\mathcal{D} = \{\mathbf{X}_i, \mathbf{Y}_i\}_{i=1}^N$, the elements of the label vector $\mathbf{Y}_i$ belong to the set $\{1,-1,?\}$, where ``1" and ``-1" represent the positive and negative labels, respectively, and ``?" denotes the missing labels. 
\begin{itemize} 
	\item \textbf{Explicit incomplete hard multi-label learning}  When the missing labels in the label vectors are explicitly marked, i.e., the label vector $\mathbf{Y}_i \in \{1,-1,?\}^C$, where ``?" explicitly records the positions of missing labels \cite{ben2022multi,kim2023bridging}. Learning from such form of data calls explicit incomplete hard multi-label learning.
	\item \textbf{Implicit incomplete hard multi-label learning}  When the missing labels in the label vectors remain unknown, i.e., the label vector $\mathbf{Y}_i \in \{1,-1\}^C$, here ``-1" denotes negative labels \textbf{or missing labels} \cite{sun2010multi,chen2023semantic}. To prevent ambiguity, it is necessary to explicitly specify the scenario as the InMLL, when the form of label vector in InMLL is the same as that in complete MLL. 
\end{itemize}

\noindent{\textbf{Incomplete soft multi-label learning}} When the elements of the label vector fall within the real number interval $[0, 1]$ with missing values, then learning from such incomplete data is termed as incomplete soft multi-label learning. Formally, given a multi-label training set $\mathcal{D} = \{\mathbf{X}_i, \mathbf{Y}_i\}_{i=1}^N$, the elements of the label vector $\mathbf{Y}_i$ belong to the set $\{a,?|a\in[0,1]\}$, where the element $\mathbf{Y}_{ij} (i \in \{1,2,\cdots,N\},j \in \{1,2,\cdots,C\})$ in the interval of $[0, 1]$ denotes the probability that the $j$-th label belongs to the $i$-th sample, while ``?" signifies the missing labels. Note that, when the labels satisfy the probability simplex constraint, the incomplete soft multi-label learning is also known as incomplete label distribution learning \cite{xu2017incomplete,li2023no}.

Since soft labels are much more difficult to be obtained in practice, existing InMLL works mainly concentrate on incomplete hard multi-label learning, while the incomplete soft multi-label learning remains relatively under-explored.

Based on the label missing type, existing InMLL works can be divided into three subcategories: weakly supervised InMLL, semi-supervised InMLL, and weakly semi-supervised InMLL.

\noindent{\textbf{Weakly supervised InMLL}} In the context of InMLL, when the provided labels for each sample in the training set is a subset of its complete labels, learning from such incomplete data is identified as weakly supervised InMLL \cite{chen2022structured,ding2023exploring,wen2023deep,liu2024masked}. Formally, given a multi-label training set $\mathcal{D} = \{\mathbf{X}_i, \mathbf{Y}_i\}_{i=1}^N$, for ease of definition, the corresponding label indicator matrix is denoted as $\mathbf{\Omega} \in \{0,1\}^{N \times C}$, where ``0" signifies the corresponding label is missing and ``1" indicates the corresponding label is given. Suppose the number of complete labels for each sample is $C_i$, $\forall i \in \{1,2,\cdots,N\}$, if $\textstyle \sum_{j=1}^{C} \mathbf{\Omega}_{ij} < C_i$ holds, then learning from such form of data is called weakly supervised InMLL, also known as multi-label learning with missing labels \cite{wang2024missing}.

\noindent{\textbf{Semi-supervised InMLL}} In the context of InMLL, when the training set includes only part of samples with complete labels (fully supervised samples), while the remaining samples have no labels provided at all (unsupervised samples), it is referred to as semi-supervised InMLL \cite{wang2021semi,li2021online,zhang2022safe,wu2023conditional}. Formally, given a multi-label training set $\mathcal{D} = \{\mathbf{X}_i, \mathbf{Y}_i\}_{i=1}^N$, for ease of definition, the set of fully supervised samples is denoted as $\Omega_L$, and the set of unsupervised samples is denoted as $\Omega_U$, then in the paradigm of semi-supervised InMLL, we have,

\begin{align}
{\mathbf{\Omega}}_{ij} =
\begin{cases}
1         & \text{if } i \in \Omega_L\\
0         & \text{if } i \in \Omega_U.
\end{cases}
\end{align}

\noindent{\textbf{Weakly semi-supervised InMLL}} In the context of InMLL, when only part of samples in the training set has a subset of complete labels provided (weakly supervised samples), while the remaining samples have no labels provided at all (unsupervised samples), it is referred to as weakly semi-supervised InMLL \cite{wu2015weakly}. Formally, given a multi-label training set $\mathcal{D} = \{\mathbf{X}_i, \mathbf{Y}_i\}_{i=1}^N$, for ease of definition, the set of weakly supervised samples is denoted as $\Omega_W$, and the set of unsupervised samples is denoted as $\Omega_U$, then in the paradigm of weakly semi-supervised InMLL, we have,

\begin{align}
\sum_{j=1}^{C} \mathbf{\Omega}_{ij}:
\begin{cases}
< C_i         & \text{if } i \in \Omega_W\\
= 0        & \text{if } i \in \Omega_U.
\end{cases}
\end{align}

Compared with weakly supervised InMLL and semi-supervised InMLL, weakly semi-supervised InMLL poses greater challenges and has received less attention in current research. Due to the relative ease of annotation, weakly semi-supervised InMLL may point out a research direction that aligns more closely with practical applications.

\subsection{Algorithm-Oriented}

From the algorithm-oriented perspective, we classify the InMLL works into four categories based on the techniques they employ: low-rank based InMLL, graph-model based InMLL, probabilistic-model based InMLL, loss design based InMLL. Figure \ref{fig2} illustrates the detailed taxonomy of InMLL.

\begin{figure*}[t]
	\centering
	\includegraphics[width=0.75\linewidth]{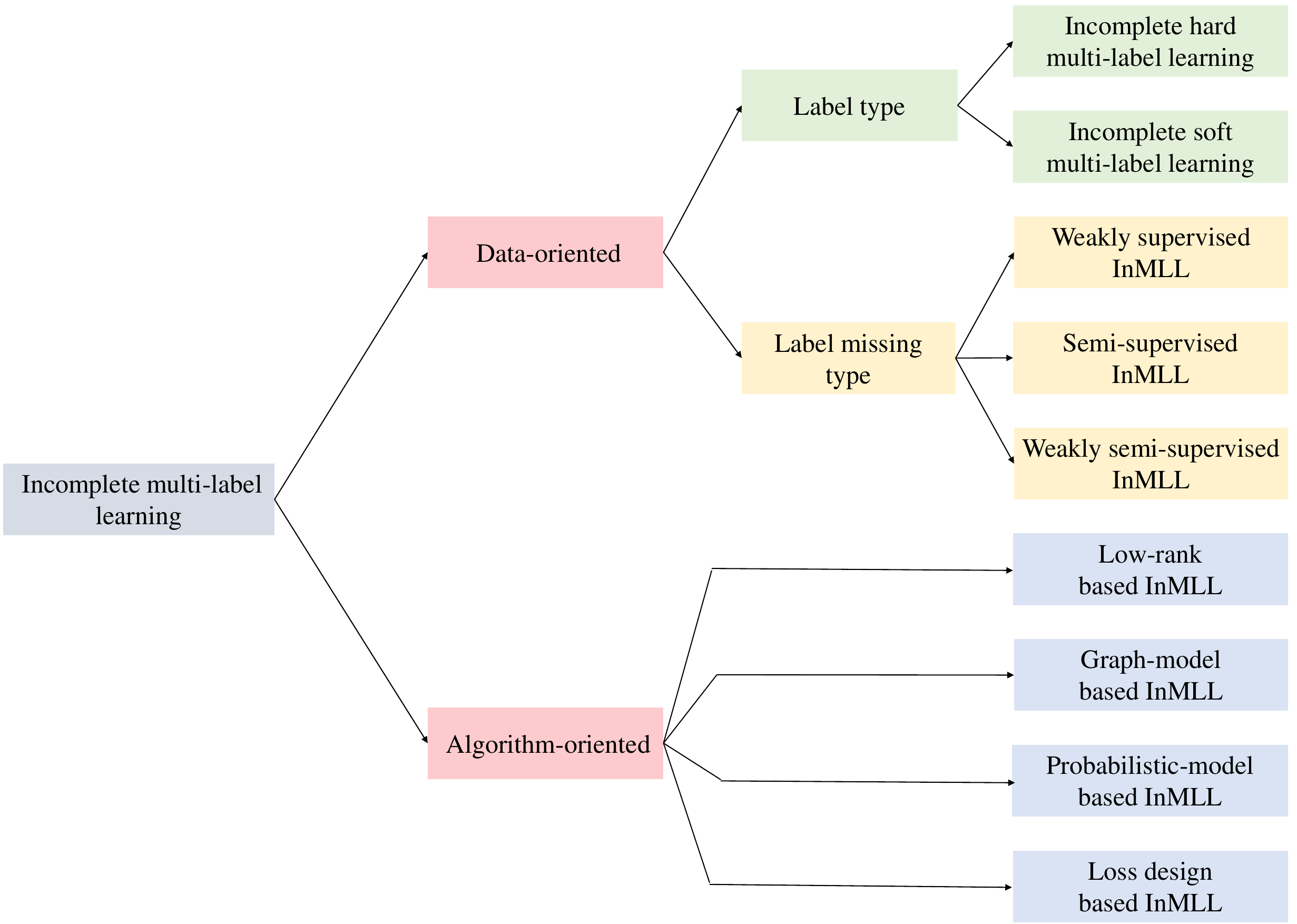} 
	\caption{A taxonomy of incomplete multi-label learning.}
	\label{fig2}
\end{figure*}

\noindent{\textbf{Low-rank based InMLL}} The low-rank assumption has been widely used in the field of matrix completion, owing to its rigorous theoretical foundation. Besides, in the field of MLL, it is often assumed that strongly correlated labels exhibit significant similarity. Consequently, the low-rank assumption is frequently employed to either complete the missing labels or characterize the correlations among labels.

Typically, low-rank based InMLL methods manifest in three distinct forms. The first kind involves directly assuming that the incomplete label matrix is a low-rank matrix, the second kind assumes that the coefficient matrix of the linear model is a low-rank matrix, and the third kind assumes that the label correlation matrix is a low-rank matrix.

In the following, we will provide the general formulations of these three kinds of methods. Let $\tilde{\mathbf{Y}}_{ij} = f^{j}\left(\mathbf{X}_{i} ; \mathbf{Z}\right)$ be the prediction of $j$-th label corresponding to the $i$-th sample, where $f$ is the prediction function and $\mathbf{Z}$ is its coefficient matrix. Then the formal objective function for the first kind of methods that directly models the incomplete label matrix as a low-rank matrix can be written as follows,

\begin{equation}
\label{eq3}
\begin{array}{c}
 \min \underset{{i, j}} \sum \mathcal{L}\left(\mathbf{Y}_{i j}, \tilde{\mathbf{Y}}_{ij}\right)+\lambda \cdot \mathit{R}(\tilde{\mathbf{Y}}), \\
\text { s.t. } \operatorname{rank}(\tilde{\mathbf{Y}}) \leq k,
\end{array}
\end{equation}
where $\mathcal{L}$ denotes the loss function, $\mathit{R}$ represent the regularization term, $\lambda$ is a trade-off hyper-parameter, and $k$ is a constant to constrain the rank. Given that minimizing rank is NP-hard, minimizing the nuclear norm is often used as an alternative.

\cite{xu2013speedup} formalize the InMLL problem as a matrix completion problem and assume that the incomplete label matrix is a low-rank matrix. Besides, they also introduce side information matrices to accelerate calculations, resulting in an algorithm with a convergence rate of $O(1/T^2)$, where $T$ is the number of iterations. \cite{han2018multi} consider the problem that both the features and labels are incomplete and cast the problem as a positive-unlabeled (PU) learning process. Based on the idea of collaborative embedding and by assuming the incomplete label matrix to be low-rank, they propose a flexible framework that is capable of jointly optimizing the incomplete features and labels.

Similarly, the formal objective function for the second kind of methods that models the coefficient matrix of the linear model as a low-rank matrix can be written as follows,

\begin{equation}
\label{eq4}
\begin{array}{c}
\min \underset{{i, j}} \sum \mathcal{L}\left(\mathbf{Y}_{i j}, f^{j}\left(\mathbf{X}_{i} ; \mathbf{Z}\right)\right)+\lambda \cdot \mathit{R}(\mathbf{Z}), \\
\text { s.t. } \operatorname{rank}(\mathbf{Z}) \leq k,
\end{array}
\end{equation}

\cite{yu2014large} study the large scale InMLL problem, to reduce the number of parameters for training, the authors assume that the coefficient matrix of the linear model is a low-rank matrix and adopt the empirical risk minimization to solve the problem. Furthermore, they provide an analysis of excess risk under the assumption that the labels are uniformly random missing. This is also the first work on theoretical generalization analysis in the field of InMLL. \cite{sun2021beyond} consider the scenario where missing and noisy labels coexist, and adopt the low-rank and sparse constraints to jointly characterize the properties of missing labels and noisy labels.

Note that, when adopting the empirical risk minimization framework with the $\ell_2$ loss function, and the low-rank regularization term is trace norm, \cite{yang2016improving} have proved that directly assuming the label matrix to be low rank is equivalent to assuming the low-rank of the coefficient matrix.

Finally, the formal objective function for the third kind of methods that models the label correlation matrix as a low-rank matrix can be written as follows,

\begin{equation}
\label{eq5}
\begin{array}{c}
\min \underset{{i, j}} \sum \mathcal{L}\left( \left( \mathbf{Y}_{i j}; \mathbf{S}\right), \tilde{\mathbf{Y}}_{ij} \right)+\lambda \cdot \mathit{R}(\mathbf{S}), \\
\text { s.t. } \operatorname{rank}(\mathbf{S}) \leq k,
\end{array}
\end{equation}
where $\mathbf{S}$ is the label correlation matrix.

\cite{tan2018incomplete} consider the problem of incomplete multi-view and weak MLL. They utilize the consistent principle of multi-view learning and the low-rank of label correlation matrix to address the incomplete views and missing labels, respectively.

In summary, due to the fact that under the low-rank assumption matrix completion is guaranteed with rigorous theoretical results, and the optimization problems are usually convex, thus global optima can be obtained. Despite these advantages, \cite{li2021concise} discover that the label matrix is just locally low-rank but globally high-rank, which reveals that the global low-rank assumptions employed in the low-rank based InMLL methods may be violated in practice.  

\noindent{\textbf{Graph-model based InMLL}} In addition to the low-rank based InMLL methods, another kind of prevalent method for mining label correlation is the graph-model based InMLL. The basic idea of this method is to conceptualize multi-label data as a graph and then apply graph learning methods to solve InMLL problems. Formally, given the multi-label training data, its corresponding graph is denoted as $\mathcal{G}=(\mathcal{V},\mathcal{E},\mathcal{W})$, where $\mathcal{V}$ is the set of vertices, $\mathcal{E}$ is the set of edges connecting the vertices, and $\mathcal{W}$ is the set of weights associated with the edges. Based on the different meanings of the vertices in constructing the graph, graph model based methods can be mainly divided into two categories: sample correlation graph based method \cite{braytee2017multi,zhu2018multi} and label correlation graph based method \cite{zhu2017multi,Wen2023Incomp}.

In the sample correlation graph, the vertices correspond to samples, i.e., $\mathcal{V}=\{\mathbf{X}_i|1 \le i \le N \}$, and $\mathbf{W}_{ij} \in \mathcal{W}$ is the weight connecting samples $i$ and $j$, usually indicating the similarity between the two samples. The degree of the $i$th sample is defined as the sum of the weights corresponding to the edges of the adjacent samples: $d_i=\sum_{j=1}^N \mathbf{W}_{ij}$, then the degree matrix can be written as $ \mathbf{D} = Diag(d_1,d_2,\cdots,d_N)$, where $Diag$ represents a diagonal matrix, and its diagonal elements are $d_1,d_2,\cdots,d_N$. Furthermore, the Laplacian matrix is defined as: $\mathbf{L}=\mathbf{D}-\mathbf{W}$. The sample correlation graph based methods posit that similar samples should have similar labels, and introduce manifold regularization to capture this underlying structure. 

Similarly, in the label correlation graph, the vertices correspond to labels, i.e., $\mathcal{V}=\{\mathbf{Y}_j|1 \le j \le C \}$. In this case, $\mathcal{W}$ is the set of weights that connecting labels, usually indicating the similarity between labels. The label correlation graph based methods believe that if two labels have similar semantics, then the label vectors containing these two categories should also be similar, and the label manifold regularization is also adopted. Some works consider both of the two correlation graphs to jointly learn the correlations of samples and labels \cite{wu2018multi,liu2018svm}.

Besides, some InMLL works employ Graph Neural Networks (GNNs) to improve label predictions. For instance, \cite{durand2019learning} develop an approach based on GNNs to explicitly model the correlation between categories and explore the curriculum learning to predict missing labels.

In summary, graph model based methods are effective in characterizing the sample correlations and label correlations. However, the computational complexity for constructing the similarity matrix is $O(N^2)$ (sample similarity) or $O(C^2)$ (label similarity), making them unable to handle large-scale data or extreme multi-label problems.

\noindent{\textbf{Probabilistic-model based InMLL}} Since the probabilistic model is capable of effectively capturing the data distribution, and the missing labels can be regarded as latent variables, making the probabilistic model based InMLL method be one of the popular methods. Among them, the commonly used Bayesian model explicitly models uncertainty using probability distributions, which enables it to address the challenge of missing labels by estimating the posterior distribution of them. Furthermore, the Bayesian model can easily incorporate prior knowledge, providing valuable assistance for the model to predict missing labels. For these reasons, the Bayesian model constitute a mainstream approach in probabilistic-model based InMLL methods. We will review several representative methods in the following.

\cite{nguyen2016bayesian} propose a Bayesian non-parametric multi-label classification model that jointly learns the latent spaces of label-feature and estimates a classifier for each label to handle the uncertainty of missing labels appropriately. By utilizing the stochastic inference, an efficient posterior inference algorithm is derived. \cite{zhao2018bayesian} present a probabilistic, fully Bayesian framework based on a generative latent factor model for learning a joint low-rank embedding of the label matrix and the label co-occurrence matrix, which is capable of effectively addressing the issue of missing labels. By leveraging both instance label vector and feature vector sparsity, a very efficient inference is obtained. 

In addition to Bayesian methods, there are some non-Bayesian methods for probabilistic-model based InMLL. For instance, \cite{jain2017scalable} propose a scalable, generative framework consisting of a latent factor model for the binary label matrix, which is coupled with an exposure model to account for label missingness. The proposed framework admits a simple inference procedure,
such that the parameter estimation reduces to a sequence of simple weighted least square regression problems.

In summary, probabilistic-model based InMLL methods can effectively capture the uncertainty of missing labels by using latent variable models and seamlessly embed prior knowledge to facilitate the learning process. Although there exist some simple non-parametric models, most of probabilistic-model based InMLL methods exhibit high computational complexity when estimating parameters and have to make specific assumptions about the data distribution.

\noindent{\textbf{Loss design based InMLL}} Binary cross entropy (BCE) loss is probably one of the most widely used losses in deep learning. Due to the symmetric design of BCE, positive and negative labels are treated equally. However, there exists severe imbalance between positives and negatives, especially in the InMLL scenario. Thus, several InMLL works resort to design new losses to discover good alternatives of BCE loss to address such imbalance issue. For instance, \cite{huang2022idea} design a self-paced correction loss, which introduces an indicator function to re-weight the positives and negatives, by such designing, pseudo positive labels would be generated to correct the direction of optimization. Besides, \cite{ben2022multi} illustrate that ignoring the un-annotated labels may lead to a limited decision boundary, while treating the un-annotated labels as negative may produce suboptimal decision boundary as it adds noisy labels. To address these issues, they introduce a class-aware selective loss to select each label individually by utilizing the label likelihood and label prior. In addition, \cite{ke2022hyperspherical} argue that the relationship between the label embeddings and image features is not symmetrical and thus can be challenging to learn in the Euclidean space. To alleviate this problem, they design a novel loss in the hyperspherical space, in which the angular margin can be incorporated into a hyperspherical multi-label loss function. This margin is effectively in balancing the impact of false negative and true positive labels. Recently, \cite{zhang2023learning} consider the long-tailed and partially labeled multi-label classification problem, and propose a novel multi-focal modifier loss that simultaneously addresses head-tail imbalance and positive-negative imbalance to modify the attention to different samples under the long-tailed class distribution.

In summary, most of loss design based InMLL methods have discovered novel losses surpassing the BCE loss, which effectively addresses the label imbalance issue. However, in designing these novel losses, additional parameters may be introduced, which adds new costs of parameter tuning.

\subsection{Single Positive Multi-Label Learning}

Single positive multi-label learning (SPMLL) is an emerging field in the InMLL, where only one positive label is annotated in each sample. The learning paradigm of SPMLL can significantly reduce the cost of annotation and align with the reality. For instance, each image of the famous ImageNet is only annotated with one single label but with multiple objects co-occur. Therefore, SPMLL has attracted a lot of attention and a plenty of SPMLL works have been proposed. SPMLL is first proposed by \cite{cole2021multi} considering its significantly reduced annotation costs. Due to the lack of precise supervision, it takes the assume negative (AN) assumption on the unobserved labels combined with various re-weighting strategies by incorporating the expected number of true positive labels, which is not available in reality. Following \cite{cole2021multi}, \cite{kim2022large} also take the AN assumption and cast the SPMLL task into noisy multi-label classification. Later, instead of making the AN assumption, \cite{zhou2022acknowledging} treats all unannotated labels as unknown by maximizing the entropy, and then adopts a heuristic asymmetric pseudo-labeling method. Notably, \cite{xu2022one} have proven that single positive label is sufficient for MLL by deriving a risk estimator approximately converging to the optimal risk minimizer of fully supervised learning, which provides a solid theoretical support. Inspired by the empirical success of consistency regularization in multi-class classification, \cite{xie2022label} extends this popular regularization to SPMLL with the help of their proposed label-aware attention module. Subsequently, \cite{chen2023semantic} present a semantic contrastive bootstrapping approach to gradually recover the cross-object relationships by introducing class activation as semantic guidance. With this learning guidance, the authors then propose a recurrent semantic masked transformer to extract iconic object-level representations and delve into the contrastive learning problems on multi-label classification tasks. \cite{liu2023revisiting} provide the conditions of effectively learning from pseudo-label for SPMLL and prove the learnability of pseudo-label based methods. Furthermore, based on the theoretical guarantee of pseudo-label for SPMLL, they propose a novel mutual label enhancement for SPMLL and prove that the generated pseudo-label by this method approximately converges to the fully-supervised case. Recently, \cite{wang2023hierarchical} propose a novel hierarchical semantic prompt network to harness the hierarchical semantic relationships using a pretrained vision and language model, i.e., Contrastive Language-Image Pretraining (CLIP), for SPMLL.

Although these methods have made great efforts and achieved promising results, there is still a large gap between the performance of SPMLL and the complete MLL, especially on data sets where the number of ground truth labels is large, such as the CUB \cite{wah2011caltech} dataset.

\begin{figure}[t]
	\centering
	\includegraphics[width=0.95\linewidth]{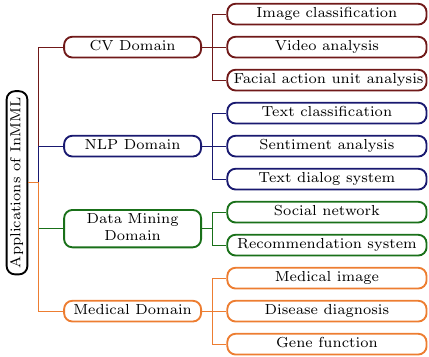} 
	\caption{Applications of incomplete multi-label learning.}
	\label{fig3}
\end{figure}

\section{Applications of InMLL} \label{sec4}

In this section, we list several applications of InMLL. Since the setting of InMLL is more aligned with the reality, it has been widely used in many fields such as computer vision (CV), natural language processing (NLP), and data mining.

In CV domain, the most common application of InMLL is image classification. Additionally, InMLL can be applied in video analysis and facial action unit analysis. 

In NLP domain, text classification is a common application of InMLL. Besides, sentiment analysis is one of the important applications, aiming to identify emotional states in text. Moreover, InMLL also has applications in text dialog system.

In data mining domain, social networks and recommendation systems are common applications. In social networks, some users may lack personal information due to the privacy protection; and in recommendation systems, users may be interested in multiple items, but the ratings of some items could be incomplete. InMLL can be applied in these situations.

Except for above applications, InMLL can be used in the medical field, such as medical image, disease diagnosis, gene function, where complete labels are usually hard to obtain. In summary, Figure \ref{fig3} shows the applications of InMLL.

\section{Future Trends} \label{sec6}

In this section, we will provide some potential future trends from two perspectives: problem setting and technical route. 
The provided trends serve as the catalyst to motivate new trends.  

From the problem setting perspective, there still remains at least four open problems. 

The first one is the \textbf{non-random missing multi-label problem}. Existing InMLL methods always assume that labels are uniformly random missing. However, this is often violated in practice \cite{arroyo2023understanding}. In real scenarios, limited by the annotator's knowledge and time, labels that are difficult to identify, require professional knowledge to annotate, and not well known are relatively more prone to be missing. For example, in medical imaging, the annotation of the rare and complicated diseases requires experts with professional knowledge. Ordinary annotators often have to give up annotating when faced with such data. A more realistic setting is that labels that are difficult to obtain are more likely to be missing. 

The second one is the \textbf{weakly semi-supervised InMLL problem}. In semi-supervised InMLL, it is usually assumed that some samples are completely labeled and others are unlabeled. In real applications, even providing complete labels of some samples still requires a large cost of resources and time. A more time-saving and labor-saving setting is that some samples only need to be annotated with a small proportion of labels, while other samples are unlabeled. This scenario requires less label information, but it undoubtedly increases the difficulty of InMLL. Its challenges mainly come from two aspects. The first is how to deal with the problem of missing labels in the incomplete labeled samples, and the second is how to utilize unlabeled samples. Note that due to the incomplete and insufficient supervision information of labeled samples, the results of the classifier trained by the guidance of such information may not be accurate, which makes the traditional semi-supervised learning algorithm not appropriate for this scenario. Therefore, it is necessary to customize algorithms for this problem, and it could also be one of the potential future research directions.

The third one is the \textbf{InMLL open-set problem}. In traditional open-set learning, it is usually assumed that samples belong to only one category, and the task is to correctly classify samples of the known categories and reject samples of unknown categories. In multi-label open set learning, a test sample may contain both known categories and unknown categories. Therefore, the task is no longer to reject samples of unknown categories, but to classify known categories in the sample and detect the unknown categories then reject them. The key of this problem is to learn how to disentangle the labels. Furthermore, in the setting of InMLL, the missing labels may hinder the learning of some labels. Therefore, how to distinguish whether the label has appeared in training but has not been provided, or whether it is an unknown label that has never appeared in training, is a new challenge in the setting of InMLL open set problem.

The fourth one is the \textbf{InMLL with noisy labels problem}. In reality, annotators often tend to label familiar targets first when faced with multi-label annotation tasks. For unfamiliar targets, they will choose to ignore these targets, guess a similar label, or even make arbitrary labels. Therefore, in real data, the issues of missing and noisy labels will arise simultaneously. Note that in some InMLL works, missing labels are usually assumed to be negative labels, which results in false negative labels as noisy labels. Here the noisy labels we emphasize referring to false positive labels. How to deal with the interference and impact of the both kinds of noisy labels with the incomplete labels is the biggest challenge in this problem.

From the technical route perspective, the three techniques below may shed new light on potential research directions.

\textbf{Multimodal large language model based techniques}. The Contrastive Language-Image Pretraining (CLIP) has received widespread attention due to its powerful capability in representation learning. CLIP was originally designed for single-label multi-class learning. However, in the InMLL scenario, each sample contains multiple labels, thus traditional contrastive loss cannot be directly used for training. How to effectively extend CLIP to the InMLL scenario is one of the directions worth exploring. At present, some works \cite{sun2022dualcoop,wang2023hierarchical,ding2023exploring} has made preliminary attempts in this direction. By leveraging multi-modal large language models, it is possible to improve the performance of InMLL to further narrow the gap between MLL and InMLL.

\textbf{Imbalanced learning based techniques}. Label imbalance is a natural and inherent problem in MLL, but it is under-explored in InMLL. Different from imbalanced learning in multi-class, the imbalance in multi-label includes two types, one is the imbalance between positive labels and negative labels, and the other is the imbalance between positive labels. Due to the coupling of multi-label, the resampling techniques commonly used in imbalanced learning are no longer suitable for multi-label learning. In the setting of InMLL, the above two imbalances will be further aggravated. Designing a new loss function to address these two imbalances is one of the potential future research directions.

\textbf{Noisy label learning based techniques}. Whether existing InMLL works adopt pseudo-labeling, label completion, or simply follow the AN assumption, they will inevitably produce noisy labels. Consequently, it is natural and reasonable to utilize noisy label learning to facilitate InMLL. Although the commonly used techniques in noisy label learning, such as label cleaning, label smoothing, and label correction, can also be explored to help design models for InMLL, the performance gap between InMLL and complete MLL is still large. Moreover, the severe label imbalance and label insufficiency problems put forward new challenges for designing algorithms that robust to noisy labels in the setting of InMLL, thus more efforts are required in this direction.

\ifCLASSOPTIONcaptionsoff
  \newpage
\fi



%
\bibliographystyle{IEEEtran}
\bibliography{IEEEabrv,InMLLSurvey}



%








\end{document}